\documentstyle[11pt,times,restrained]{article}
\include{psfig}
\newcounter{treecount}
\newcounter{branchcount}
\newsavebox{\parentbox}
\newsavebox{\treebox}
\newsavebox{\treeboxone}
\newsavebox{\treeboxtwo}
\newsavebox{\treeboxthree}
\newsavebox{\treeboxfour}
\newsavebox{\treeboxfive}
\newsavebox{\treeboxsix}
\newsavebox{\treeboxseven}
\newsavebox{\treeboxeight}
\newsavebox{\treeboxnine}
\newsavebox{\treeboxten}
\newsavebox{\treeboxeleven}
\newsavebox{\treeboxtwelve}
\newsavebox{\treeboxthirteen}
\newsavebox{\treeboxfourteen}
\newsavebox{\treeboxfifteen}
\newsavebox{\treeboxsixteen}
\newsavebox{\treeboxseventeen}
\newsavebox{\treeboxeighteen}
\newsavebox{\treeboxnineteen}
\newsavebox{\treeboxtwenty}
\newlength{\treeoffsetone}
\newlength{\treeoffsettwo}
\newlength{\treeoffsetthree}
\newlength{\treeoffsetfour}
\newlength{\treeoffsetfive}
\newlength{\treeoffsetsix}
\newlength{\treeoffsetseven}
\newlength{\treeoffseteight}
\newlength{\treeoffsetnine}
\newlength{\treeoffsetten}
\newlength{\treeoffseteleven}
\newlength{\treeoffsettwelve}
\newlength{\treeoffsetthirteen}
\newlength{\treeoffsetfourteen}
\newlength{\treeoffsetfifteen}
\newlength{\treeoffsetsixteen}
\newlength{\treeoffsetseventeen}
\newlength{\treeoffseteighteen}
\newlength{\treeoffsetnineteen}
\newlength{\treeoffsettwenty}

\newlength{\treeshiftone}
\newlength{\treeshifttwo}
\newlength{\treeshiftthree}
\newlength{\treeshiftfour}
\newlength{\treeshiftfive}
\newlength{\treeshiftsix}
\newlength{\treeshiftseven}
\newlength{\treeshifteight}
\newlength{\treeshiftnine}
\newlength{\treeshiftten}
\newlength{\treeshifteleven}
\newlength{\treeshifttwelve}
\newlength{\treeshiftthirteen}
\newlength{\treeshiftfourteen}
\newlength{\treeshiftfifteen}
\newlength{\treeshiftsixteen}
\newlength{\treeshiftseventeen}
\newlength{\treeshifteighteen}
\newlength{\treeshiftnineteen}
\newlength{\treeshifttwenty}
\newlength{\treewidthone}
\newlength{\treewidthtwo}
\newlength{\treewidththree}
\newlength{\treewidthfour}
\newlength{\treewidthfive}
\newlength{\treewidthsix}
\newlength{\treewidthseven}
\newlength{\treewidtheight}
\newlength{\treewidthnine}
\newlength{\treewidthten}
\newlength{\treewidtheleven}
\newlength{\treewidthtwelve}
\newlength{\treewidththirteen}
\newlength{\treewidthfourteen}
\newlength{\treewidthfifteen}
\newlength{\treewidthsixteen}
\newlength{\treewidthseventeen}
\newlength{\treewidtheighteen}
\newlength{\treewidthnineteen}
\newlength{\treewidthtwenty}
\newlength{\daughteroffsetone}
\newlength{\daughteroffsettwo}
\newlength{\daughteroffsetthree}
\newlength{\daughteroffsetfour}
\newlength{\branchwidthone}
\newlength{\branchwidthtwo}
\newlength{\branchwidththree}
\newlength{\branchwidthfour}
\newlength{\parentoffset}
\newlength{\treeoffset}
\newlength{\daughteroffset}
\newlength{\branchwidth}
\newlength{\parentwidth}
\newlength{\treewidth}
\newcommand{\ontop}[1]{\begin{tabular}{c}#1\end{tabular}}
\newcommand{\poptree}{%
\ifnum\value{treecount}=0\typeout{QobiTeX warning---Tree stack underflow}\fi%
\addtocounter{treecount}{-1}%
\setlength{\treeoffsettwo}{\treeoffsetthree}%
\setlength{\treeoffsetthree}{\treeoffsetfour}%
\setlength{\treeoffsetfour}{\treeoffsetfive}%
\setlength{\treeoffsetfive}{\treeoffsetsix}%
\setlength{\treeoffsetsix}{\treeoffsetseven}%
\setlength{\treeoffsetseven}{\treeoffseteight}%
\setlength{\treeoffseteight}{\treeoffsetnine}%
\setlength{\treeoffsetnine}{\treeoffsetten}%
\setlength{\treeoffsetten}{\treeoffseteleven}%
\setlength{\treeoffseteleven}{\treeoffsettwelve}%
\setlength{\treeoffsettwelve}{\treeoffsetthirteen}%
\setlength{\treeoffsetthirteen}{\treeoffsetfourteen}%
\setlength{\treeoffsetfourteen}{\treeoffsetfifteen}%
\setlength{\treeoffsetfifteen}{\treeoffsetsixteen}%
\setlength{\treeoffsetsixteen}{\treeoffsetseventeen}%
\setlength{\treeoffsetseventeen}{\treeoffseteighteen}%
\setlength{\treeoffseteighteen}{\treeoffsetnineteen}%
\setlength{\treeoffsetnineteen}{\treeoffsettwenty}%
\setlength{\treeshifttwo}{\treeshiftthree}%
\setlength{\treeshiftthree}{\treeshiftfour}%
\setlength{\treeshiftfour}{\treeshiftfive}%
\setlength{\treeshiftfive}{\treeshiftsix}%
\setlength{\treeshiftsix}{\treeshiftseven}%
\setlength{\treeshiftseven}{\treeshifteight}%
\setlength{\treeshifteight}{\treeshiftnine}%
\setlength{\treeshiftnine}{\treeshiftten}%
\setlength{\treeshiftten}{\treeshifteleven}%
\setlength{\treeshifteleven}{\treeshifttwelve}%
\setlength{\treeshifttwelve}{\treeshiftthirteen}%
\setlength{\treeshiftthirteen}{\treeshiftfourteen}%
\setlength{\treeshiftfourteen}{\treeshiftfifteen}%
\setlength{\treeshiftfifteen}{\treeshiftsixteen}%
\setlength{\treeshiftsixteen}{\treeshiftseventeen}%
\setlength{\treeshiftseventeen}{\treeshifteighteen}%
\setlength{\treeshifteighteen}{\treeshiftnineteen}%
\setlength{\treeshiftnineteen}{\treeshifttwenty}%
\setlength{\treewidthtwo}{\treewidththree}%
\setlength{\treewidththree}{\treewidthfour}%
\setlength{\treewidthfour}{\treewidthfive}%
\setlength{\treewidthfive}{\treewidthsix}%
\setlength{\treewidthsix}{\treewidthseven}%
\setlength{\treewidthseven}{\treewidtheight}%
\setlength{\treewidtheight}{\treewidthnine}%
\setlength{\treewidthnine}{\treewidthten}%
\setlength{\treewidthten}{\treewidtheleven}%
\setlength{\treewidtheleven}{\treewidthtwelve}%
\setlength{\treewidthtwelve}{\treewidththirteen}%
\setlength{\treewidththirteen}{\treewidthfourteen}%
\setlength{\treewidthfourteen}{\treewidthfifteen}%
\setlength{\treewidthfifteen}{\treewidthsixteen}%
\setlength{\treewidthsixteen}{\treewidthseventeen}%
\setlength{\treewidthseventeen}{\treewidtheighteen}%
\setlength{\treewidtheighteen}{\treewidthnineteen}%
\setlength{\treewidthnineteen}{\treewidthtwenty}%
\sbox{\treeboxtwo}{\usebox{\treeboxthree}}%
\sbox{\treeboxthree}{\usebox{\treeboxfour}}%
\sbox{\treeboxfour}{\usebox{\treeboxfive}}%
\sbox{\treeboxfive}{\usebox{\treeboxsix}}%
\sbox{\treeboxsix}{\usebox{\treeboxseven}}%
\sbox{\treeboxseven}{\usebox{\treeboxeight}}%
\sbox{\treeboxeight}{\usebox{\treeboxnine}}%
\sbox{\treeboxnine}{\usebox{\treeboxten}}%
\sbox{\treeboxten}{\usebox{\treeboxeleven}}%
\sbox{\treeboxeleven}{\usebox{\treeboxtwelve}}%
\sbox{\treeboxtwelve}{\usebox{\treeboxthirteen}}%
\sbox{\treeboxthirteen}{\usebox{\treeboxfourteen}}%
\sbox{\treeboxfourteen}{\usebox{\treeboxfifteen}}%
\sbox{\treeboxfifteen}{\usebox{\treeboxsixteen}}%
\sbox{\treeboxsixteen}{\usebox{\treeboxseventeen}}%
\sbox{\treeboxseventeen}{\usebox{\treeboxeighteen}}%
\sbox{\treeboxeighteen}{\usebox{\treeboxnineteen}}%
\sbox{\treeboxnineteen}{\usebox{\treeboxtwenty}}}
\newcommand{\leaf}[1]{%
\ifnum\value{treecount}=20\typeout{QobiTeX warning---Tree stack overflow}\fi%
\addtocounter{treecount}{1}%
\sbox{\treeboxtwenty}{\usebox{\treeboxnineteen}}%
\sbox{\treeboxnineteen}{\usebox{\treeboxeighteen}}%
\sbox{\treeboxeighteen}{\usebox{\treeboxseventeen}}%
\sbox{\treeboxseventeen}{\usebox{\treeboxsixteen}}%
\sbox{\treeboxsixteen}{\usebox{\treeboxfifteen}}%
\sbox{\treeboxfifteen}{\usebox{\treeboxfourteen}}%
\sbox{\treeboxfourteen}{\usebox{\treeboxthirteen}}%
\sbox{\treeboxthirteen}{\usebox{\treeboxtwelve}}%
\sbox{\treeboxtwelve}{\usebox{\treeboxeleven}}%
\sbox{\treeboxeleven}{\usebox{\treeboxten}}%
\sbox{\treeboxten}{\usebox{\treeboxnine}}%
\sbox{\treeboxnine}{\usebox{\treeboxeight}}%
\sbox{\treeboxeight}{\usebox{\treeboxseven}}%
\sbox{\treeboxseven}{\usebox{\treeboxsix}}%
\sbox{\treeboxsix}{\usebox{\treeboxfive}}%
\sbox{\treeboxfive}{\usebox{\treeboxfour}}%
\sbox{\treeboxfour}{\usebox{\treeboxthree}}%
\sbox{\treeboxthree}{\usebox{\treeboxtwo}}%
\sbox{\treeboxtwo}{\usebox{\treeboxone}}%
\sbox{\treeboxone}{\ontop{#1}}%
\sbox{\treeboxone}{\raisebox{-\ht\treeboxone}{\usebox{\treeboxone}}}%
\setlength{\treeoffsettwenty}{\treeoffsetnineteen}%
\setlength{\treeoffsetnineteen}{\treeoffseteighteen}%
\setlength{\treeoffseteighteen}{\treeoffsetseventeen}%
\setlength{\treeoffsetseventeen}{\treeoffsetsixteen}%
\setlength{\treeoffsetsixteen}{\treeoffsetfifteen}%
\setlength{\treeoffsetfifteen}{\treeoffsetfourteen}%
\setlength{\treeoffsetfourteen}{\treeoffsetthirteen}%
\setlength{\treeoffsetthirteen}{\treeoffsettwelve}%
\setlength{\treeoffsettwelve}{\treeoffseteleven}%
\setlength{\treeoffseteleven}{\treeoffsetten}%
\setlength{\treeoffsetten}{\treeoffsetnine}%
\setlength{\treeoffsetnine}{\treeoffseteight}%
\setlength{\treeoffseteight}{\treeoffsetseven}%
\setlength{\treeoffsetseven}{\treeoffsetsix}%
\setlength{\treeoffsetsix}{\treeoffsetfive}%
\setlength{\treeoffsetfive}{\treeoffsetfour}%
\setlength{\treeoffsetfour}{\treeoffsetthree}%
\setlength{\treeoffsetthree}{\treeoffsettwo}%
\setlength{\treeoffsettwo}{\treeoffsetone}%
\setlength{\treeoffsetone}{0.5\wd\treeboxone}%
\setlength{\treeshifttwenty}{\treeshiftnineteen}%
\setlength{\treeshiftnineteen}{\treeshifteighteen}%
\setlength{\treeshifteighteen}{\treeshiftseventeen}%
\setlength{\treeshiftseventeen}{\treeshiftsixteen}%
\setlength{\treeshiftsixteen}{\treeshiftfifteen}%
\setlength{\treeshiftfifteen}{\treeshiftfourteen}%
\setlength{\treeshiftfourteen}{\treeshiftthirteen}%
\setlength{\treeshiftthirteen}{\treeshifttwelve}%
\setlength{\treeshifttwelve}{\treeshifteleven}%
\setlength{\treeshifteleven}{\treeshiftten}%
\setlength{\treeshiftten}{\treeshiftnine}%
\setlength{\treeshiftnine}{\treeshifteight}%
\setlength{\treeshifteight}{\treeshiftseven}%
\setlength{\treeshiftseven}{\treeshiftsix}%
\setlength{\treeshiftsix}{\treeshiftfive}%
\setlength{\treeshiftfive}{\treeshiftfour}%
\setlength{\treeshiftfour}{\treeshiftthree}%
\setlength{\treeshiftthree}{\treeshifttwo}%
\setlength{\treeshifttwo}{\treeshiftone}%
\setlength{\treeshiftone}{0pt}%
\setlength{\treewidthtwenty}{\treewidthnineteen}%
\setlength{\treewidthnineteen}{\treewidtheighteen}%
\setlength{\treewidtheighteen}{\treewidthseventeen}%
\setlength{\treewidthseventeen}{\treewidthsixteen}%
\setlength{\treewidthsixteen}{\treewidthfifteen}%
\setlength{\treewidthfifteen}{\treewidthfourteen}%
\setlength{\treewidthfourteen}{\treewidththirteen}%
\setlength{\treewidththirteen}{\treewidthtwelve}%
\setlength{\treewidthtwelve}{\treewidtheleven}%
\setlength{\treewidtheleven}{\treewidthten}%
\setlength{\treewidthten}{\treewidthnine}%
\setlength{\treewidthnine}{\treewidtheight}%
\setlength{\treewidtheight}{\treewidthseven}%
\setlength{\treewidthseven}{\treewidthsix}%
\setlength{\treewidthsix}{\treewidthfive}%
\setlength{\treewidthfive}{\treewidthfour}%
\setlength{\treewidthfour}{\treewidththree}%
\setlength{\treewidththree}{\treewidthtwo}%
\setlength{\treewidthtwo}{\treewidthone}%
\setlength{\treewidthone}{\wd\treeboxone}}
\newcommand{\branch}[2]{%
\setcounter{branchcount}{#1}%
\ifnum\value{branchcount}=1\sbox{\parentbox}{\ontop{#2}}%
\setlength{\parentoffset}{\treeoffsetone}%
\addtolength{\parentoffset}{-0.5\wd\parentbox}%
\setlength{\daughteroffset}{0in}%
\ifdim\parentoffset<0in%
\setlength{\daughteroffset}{-\parentoffset}%
\setlength{\parentoffset}{0in}\fi%
\setlength{\parentwidth}{\parentoffset}%
\addtolength{\parentwidth}{\wd\parentbox}%
\setlength{\treeoffset}{\daughteroffset}%
\addtolength{\treeoffset}{\treeoffsetone}%
\setlength{\treewidth}{\wd\treeboxone}%
\addtolength{\treewidth}{\daughteroffset}%
\ifdim\treewidth<\parentwidth\setlength{\treewidth}{\parentwidth}\fi%
\sbox{\treebox}{\begin{minipage}{\treewidth}%
\begin{flushleft}%
\hspace*{\parentoffset}\usebox{\parentbox}\\
{\setlength{\unitlength}{2ex}%
\hspace*{\treeoffset}\begin{picture}(0,1)%
\put(0,0){\line(0,1){1}}%
\end{picture}}\\
\vspace{-\baselineskip}
\hspace*{\daughteroffset}%
\raisebox{-\ht\treeboxone}{\usebox{\treeboxone}}%
\end{flushleft}%
\end{minipage}}%
\setlength{\treeoffsetone}{\parentoffset}%
\addtolength{\treeoffsetone}{0.5\wd\parentbox}%
\setlength{\treeshiftone}{0pt}%
\setlength{\treewidthone}{\treewidth}%
\sbox{\treeboxone}{\usebox{\treebox}}%
\else\ifnum\value{branchcount}=2\sbox{\parentbox}{\ontop{#2}}%
\setlength{\branchwidthone}{\treewidthtwo}%
\addtolength{\branchwidthone}{\treeoffsetone}%
\addtolength{\branchwidthone}{-\treeshiftone}%
\addtolength{\branchwidthone}{-\treeoffsettwo}%
\setlength{\branchwidth}{\branchwidthone}%
\setlength{\daughteroffsetone}{\branchwidth}%
\addtolength{\daughteroffsetone}{-\branchwidthone}%
\addtolength{\daughteroffsetone}{-\treeshiftone}%
\setlength{\parentoffset}{-0.5\wd\parentbox}%
\addtolength{\parentoffset}{\treeoffsettwo}%
\addtolength{\parentoffset}{0.5\branchwidth}%
\setlength{\daughteroffset}{0in}%
\ifdim\parentoffset<0in%
\setlength{\daughteroffset}{-\parentoffset}%
\setlength{\parentoffset}{0in}\fi%
\setlength{\parentwidth}{\parentoffset}%
\addtolength{\parentwidth}{\wd\parentbox}%
\setlength{\treeoffset}{\daughteroffset}%
\addtolength{\treeoffset}{\treeoffsettwo}%
\setlength{\treewidth}{\wd\treeboxone}%
\addtolength{\treewidth}{\daughteroffsetone}%
\addtolength{\treewidth}{\treewidthtwo}%
\addtolength{\treewidth}{\daughteroffset}%
\ifdim\treewidth<\parentwidth\setlength{\treewidth}{\parentwidth}\fi%
\sbox{\treebox}{\begin{minipage}{\treewidth}%
\begin{flushleft}%
\hspace*{\parentoffset}\usebox{\parentbox}\\
{\setlength{\unitlength}{0.5\branchwidth}%
\hspace*{\treeoffset}\begin{picture}(2,0.5)%
\put(0,0){\line(2,1){1}}%
\put(2,0){\line(-2,1){1}}%
\end{picture}}\\
\vspace{-\baselineskip}
\hspace*{\daughteroffset}%
\makebox[\treewidthtwo][l]%
{\raisebox{-\ht\treeboxtwo}{\usebox{\treeboxtwo}}}%
\hspace*{\daughteroffsetone}%
\raisebox{-\ht\treeboxone}{\usebox{\treeboxone}}%
\end{flushleft}%
\end{minipage}}%
\setlength{\treeoffsetone}{\parentoffset}%
\addtolength{\treeoffsetone}{0.5\wd\parentbox}%
\setlength{\treeshiftone}{0pt}%
\setlength{\treewidthone}{\treewidth}%
\sbox{\treeboxone}{\usebox{\treebox}}\poptree%
\else\ifnum\value{branchcount}=3\sbox{\parentbox}{\ontop{#2}}%
\setlength{\branchwidthone}{\treewidthtwo}%
\addtolength{\branchwidthone}{\treeoffsetone}%
\addtolength{\branchwidthone}{-\treeshiftone}%
\addtolength{\branchwidthone}{-\treeoffsettwo}%
\setlength{\branchwidthtwo}{\treewidththree}%
\addtolength{\branchwidthtwo}{\treeoffsettwo}%
\addtolength{\branchwidthtwo}{-\treeshifttwo}%
\addtolength{\branchwidthtwo}{-\treeoffsetthree}%
\setlength{\branchwidth}{\branchwidthone}%
\ifdim\branchwidthtwo>\branchwidth%
\setlength{\branchwidth}{\branchwidthtwo}\fi%
\setlength{\daughteroffsetone}{\branchwidth}%
\addtolength{\daughteroffsetone}{-\branchwidthone}%
\addtolength{\daughteroffsetone}{-\treeshiftone}%
\setlength{\daughteroffsettwo}{\branchwidth}%
\addtolength{\daughteroffsettwo}{-\branchwidthtwo}%
\addtolength{\daughteroffsettwo}{-\treeshifttwo}%
\setlength{\parentoffset}{-0.5\wd\parentbox}%
\addtolength{\parentoffset}{\treeoffsetthree}%
\addtolength{\parentoffset}{\branchwidth}%
\setlength{\daughteroffset}{0in}%
\ifdim\parentoffset<0in%
\setlength{\daughteroffset}{-\parentoffset}%
\setlength{\parentoffset}{0in}\fi%
\setlength{\parentwidth}{\parentoffset}%
\addtolength{\parentwidth}{\wd\parentbox}%
\setlength{\treeoffset}{\daughteroffset}%
\addtolength{\treeoffset}{\treeoffsetthree}%
\setlength{\treewidth}{\wd\treeboxone}%
\addtolength{\treewidth}{\daughteroffsetone}%
\addtolength{\treewidth}{\treewidthtwo}%
\addtolength{\treewidth}{\daughteroffsettwo}%
\addtolength{\treewidth}{\treewidththree}%
\addtolength{\treewidth}{\daughteroffset}%
\ifdim\treewidth<\parentwidth\setlength{\treewidth}{\parentwidth}\fi%
\sbox{\treebox}{\begin{minipage}{\treewidth}%
\begin{flushleft}%
\hspace*{\parentoffset}\usebox{\parentbox}\\
{\setlength{\unitlength}{0.5\branchwidth}%
\hspace*{\treeoffset}\begin{picture}(4,1)%
\put(0,0){\line(2,1){2}}%
\put(2,0){\line(0,1){1}}%
\put(4,0){\line(-2,1){2}}%
\end{picture}}\\
\vspace{-\baselineskip}
\hspace*{\daughteroffset}%
\makebox[\treewidththree][l]%
{\raisebox{-\ht\treeboxthree}{\usebox{\treeboxthree}}}%
\hspace*{\daughteroffsettwo}%
\makebox[\treewidthtwo][l]%
{\raisebox{-\ht\treeboxtwo}{\usebox{\treeboxtwo}}}%
\hspace*{\daughteroffsetone}%
\raisebox{-\ht\treeboxone}{\usebox{\treeboxone}}%
\end{flushleft}%
\end{minipage}}%
\setlength{\treeoffsetone}{\parentoffset}%
\addtolength{\treeoffsetone}{0.5\wd\parentbox}%
\setlength{\treeshiftone}{0pt}%
\setlength{\treewidthone}{\treewidth}%
\sbox{\treeboxone}{\usebox{\treebox}}\poptree\poptree%
\else\ifnum\value{branchcount}=4\sbox{\parentbox}{\ontop{#2}}%
\setlength{\branchwidthone}{\treewidthtwo}%
\addtolength{\branchwidthone}{\treeoffsetone}%
\addtolength{\branchwidthone}{-\treeshiftone}%
\addtolength{\branchwidthone}{-\treeoffsettwo}%
\setlength{\branchwidthtwo}{\treewidththree}%
\addtolength{\branchwidthtwo}{\treeoffsettwo}%
\addtolength{\branchwidthtwo}{-\treeshifttwo}%
\addtolength{\branchwidthtwo}{-\treeoffsetthree}%
\setlength{\branchwidththree}{\treewidthfour}%
\addtolength{\branchwidththree}{\treeoffsetthree}%
\addtolength{\branchwidththree}{-\treeshiftthree}%
\addtolength{\branchwidththree}{-\treeoffsetfour}%
\setlength{\branchwidth}{\branchwidthone}%
\ifdim\branchwidthtwo>\branchwidth%
\setlength{\branchwidth}{\branchwidthtwo}\fi%
\ifdim\branchwidththree>\branchwidth%
\setlength{\branchwidth}{\branchwidththree}\fi%
\setlength{\daughteroffsetone}{\branchwidth}%
\addtolength{\daughteroffsetone}{-\branchwidthone}%
\addtolength{\daughteroffsetone}{-\treeshiftone}%
\setlength{\daughteroffsettwo}{\branchwidth}%
\addtolength{\daughteroffsettwo}{-\branchwidthtwo}%
\addtolength{\daughteroffsettwo}{-\treeshifttwo}%
\setlength{\daughteroffsetthree}{\branchwidth}%
\addtolength{\daughteroffsetthree}{-\branchwidththree}%
\addtolength{\daughteroffsetthree}{-\treeshiftthree}%
\setlength{\parentoffset}{-0.5\wd\parentbox}%
\addtolength{\parentoffset}{\treeoffsetfour}%
\addtolength{\parentoffset}{1.5\branchwidth}%
\setlength{\daughteroffset}{0in}%
\ifdim\parentoffset<0in%
\setlength{\daughteroffset}{-\parentoffset}%
\setlength{\parentoffset}{0in}\fi%
\setlength{\parentwidth}{\parentoffset}%
\addtolength{\parentwidth}{\wd\parentbox}%
\setlength{\treeoffset}{\daughteroffset}%
\addtolength{\treeoffset}{\treeoffsetfour}%
\setlength{\treewidth}{\wd\treeboxone}%
\addtolength{\treewidth}{\daughteroffsetone}%
\addtolength{\treewidth}{\treewidthtwo}%
\addtolength{\treewidth}{\daughteroffsettwo}%
\addtolength{\treewidth}{\treewidththree}%
\addtolength{\treewidth}{\daughteroffsetthree}%
\addtolength{\treewidth}{\treewidthfour}%
\addtolength{\treewidth}{\daughteroffset}%
\ifdim\treewidth<\parentwidth\setlength{\treewidth}{\parentwidth}\fi%
\sbox{\treebox}{\begin{minipage}{\treewidth}%
\begin{flushleft}%
\hspace*{\parentoffset}\usebox{\parentbox}\\
{\setlength{\unitlength}{0.5\branchwidth}%
\hspace*{\treeoffset}\begin{picture}(6,1)%
\put(0,0){\line(3,1){3}}%
\put(2,0){\line(1,1){1}}%
\put(4,0){\line(-1,1){1}}%
\put(6,0){\line(-3,1){3}}%
\end{picture}}\\
\vspace{-\baselineskip}
\hspace*{\daughteroffset}%
\makebox[\treewidthfour][l]%
{\raisebox{-\ht\treeboxfour}{\usebox{\treeboxfour}}}%
\hspace*{\daughteroffsetthree}%
\makebox[\treewidththree][l]%
{\raisebox{-\ht\treeboxthree}{\usebox{\treeboxthree}}}%
\hspace*{\daughteroffsettwo}%
\makebox[\treewidthtwo][l]%
{\raisebox{-\ht\treeboxtwo}{\usebox{\treeboxtwo}}}%
\hspace*{\daughteroffsetone}%
\raisebox{-\ht\treeboxone}{\usebox{\treeboxone}}%
\end{flushleft}%
\end{minipage}}%
\setlength{\treeoffsetone}{\parentoffset}%
\addtolength{\treeoffsetone}{0.5\wd\parentbox}%
\setlength{\treeshiftone}{0pt}%
\setlength{\treewidthone}{\treewidth}%
\sbox{\treeboxone}{\usebox{\treebox}}\poptree\poptree\poptree%
\else\ifnum\value{branchcount}=5\sbox{\parentbox}{\ontop{#2}}%
\setlength{\branchwidthone}{\treewidthtwo}%
\addtolength{\branchwidthone}{\treeoffsetone}%
\addtolength{\branchwidthone}{-\treeshiftone}%
\addtolength{\branchwidthone}{-\treeoffsettwo}%
\setlength{\branchwidthtwo}{\treewidththree}%
\addtolength{\branchwidthtwo}{\treeoffsettwo}%
\addtolength{\branchwidthtwo}{-\treeshifttwo}%
\addtolength{\branchwidthtwo}{-\treeoffsetthree}%
\setlength{\branchwidththree}{\treewidthfour}%
\addtolength{\branchwidththree}{\treeoffsetthree}%
\addtolength{\branchwidththree}{-\treeshiftthree}%
\addtolength{\branchwidththree}{-\treeoffsetfour}%
\setlength{\branchwidthfour}{\treewidthfive}%
\addtolength{\branchwidthfour}{\treeoffsetfour}%
\addtolength{\branchwidthfour}{-\treeshiftfour}%
\addtolength{\branchwidthfour}{-\treeoffsetfive}%
\setlength{\branchwidth}{\branchwidthone}%
\ifdim\branchwidthtwo>\branchwidth%
\setlength{\branchwidth}{\branchwidthtwo}\fi%
\ifdim\branchwidththree>\branchwidth%
\setlength{\branchwidth}{\branchwidththree}\fi%
\ifdim\branchwidthfour>\branchwidth%
\setlength{\branchwidth}{\branchwidthfour}\fi%
\setlength{\daughteroffsetone}{\branchwidth}%
\addtolength{\daughteroffsetone}{-\branchwidthone}%
\addtolength{\daughteroffsetone}{-\treeshiftone}%
\setlength{\daughteroffsettwo}{\branchwidth}%
\addtolength{\daughteroffsettwo}{-\branchwidthtwo}%
\addtolength{\daughteroffsettwo}{-\treeshifttwo}%
\setlength{\daughteroffsetthree}{\branchwidth}%
\addtolength{\daughteroffsetthree}{-\branchwidththree}%
\addtolength{\daughteroffsetthree}{-\treeshiftthree}%
\setlength{\daughteroffsetfour}{\branchwidth}%
\addtolength{\daughteroffsetfour}{-\branchwidthfour}%
\addtolength{\daughteroffsetfour}{-\treeshiftfour}%
\setlength{\parentoffset}{-0.5\wd\parentbox}%
\addtolength{\parentoffset}{\treeoffsetfive}%
\addtolength{\parentoffset}{2\branchwidth}%
\setlength{\daughteroffset}{0in}%
\ifdim\parentoffset<0in%
\setlength{\daughteroffset}{-\parentoffset}%
\setlength{\parentoffset}{0in}\fi%
\setlength{\parentwidth}{\parentoffset}%
\addtolength{\parentwidth}{\wd\parentbox}%
\setlength{\treeoffset}{\daughteroffset}%
\addtolength{\treeoffset}{\treeoffsetfive}%
\setlength{\treewidth}{\wd\treeboxone}%
\addtolength{\treewidth}{\daughteroffsetone}%
\addtolength{\treewidth}{\treewidthtwo}%
\addtolength{\treewidth}{\daughteroffsettwo}%
\addtolength{\treewidth}{\treewidththree}%
\addtolength{\treewidth}{\daughteroffsetthree}%
\addtolength{\treewidth}{\treewidthfour}%
\addtolength{\treewidth}{\daughteroffsetfour}%
\addtolength{\treewidth}{\treewidthfive}%
\addtolength{\treewidth}{\daughteroffset}%
\ifdim\treewidth<\parentwidth\setlength{\treewidth}{\parentwidth}\fi%
\sbox{\treebox}{\begin{minipage}{\treewidth}%
\begin{flushleft}%
\hspace*{\parentoffset}\usebox{\parentbox}\\
{\setlength{\unitlength}{0.5\branchwidth}%
\hspace*{\treeoffset}\begin{picture}(8,1)%
\put(0,0){\line(4,1){4}}%
\put(2,0){\line(2,1){2}}%
\put(4,0){\line(0,1){1}}%
\put(6,0){\line(-2,1){2}}%
\put(8,0){\line(-4,1){4}}%
\end{picture}}\\
\vspace{-\baselineskip}
\hspace*{\daughteroffset}%
\makebox[\treewidthfive][l]%
{\raisebox{-\ht\treeboxfour}{\usebox{\treeboxfive}}}%
\hspace*{\daughteroffsetfour}%
\makebox[\treewidthfour][l]%
{\raisebox{-\ht\treeboxfour}{\usebox{\treeboxfour}}}%
\hspace*{\daughteroffsetthree}%
\makebox[\treewidththree][l]%
{\raisebox{-\ht\treeboxthree}{\usebox{\treeboxthree}}}%
\hspace*{\daughteroffsettwo}%
\makebox[\treewidthtwo][l]%
{\raisebox{-\ht\treeboxtwo}{\usebox{\treeboxtwo}}}%
\hspace*{\daughteroffsetone}%
\raisebox{-\ht\treeboxone}{\usebox{\treeboxone}}%
\end{flushleft}%
\end{minipage}}%
\setlength{\treeoffsetone}{\parentoffset}%
\addtolength{\treeoffsetone}{0.5\wd\parentbox}%
\setlength{\treeshiftone}{0pt}%
\setlength{\treewidthone}{\treewidth}%
\sbox{\treeboxone}{\usebox{\treebox}}\poptree\poptree\poptree\poptree%
\else\typeout{QobiTeX warning--- Can't handle #1 branching}\fi\fi\fi\fi\fi}
\newcommand{\tree}{%
\usebox{\treeboxone}
\setlength{\treeoffsetone}{\treeoffsettwo}%
\sbox{\treeboxone}{\usebox{\treeboxtwo}}%
\poptree}

\evensidemargin \oddsidemargin
\newcommand{\speaker}{[\mbox{\sc s}]}
\newcommand{\shared}{[\mbox{\sc c}]}
\newcommand{\distractors}{\mbox{\em D}}
\newcommand{\entity}{\mbox{\em e}}
\newcommand{\entityA}{\mbox{\em a}}
\newcommand{\entityB}{\mbox{\em b}}
\newcommand{\entities}{\mbox{\bf e}}
\newcommand{\entitiesA}{\mbox{\bf a}}
\newcommand{\variables}{\mbox{\bf v}}
\newcommand{\entitiesI}{\mbox{\bf e}_{\mbox{\footnotesize\em i}}}
\newcommand{\entitiesIA}{\mbox{\bf a}_{\mbox{\footnotesize\em i}}}
\newcommand{\variablesI}{\mbox{\bf v}_{\mbox{\footnotesize\em i}}}
\newcommand{\contributions}{\mbox{\em N}}
\newcommand{\requirements}{\mbox{\em R}}

\newcommand{\contributionsI}{\mbox{\em N}_{\mbox{\footnotesize\em i}}}
\newcommand{\requirementsI}{\mbox{\em R}_{\mbox{\footnotesize\em i}}}
\newcommand{\commgoal}{\mbox{\em G}}
\newcommand{\imp}{\supset}

\newenvironment{egs}[1]{\refstepcounter{equation}\label{#1}\begin{list}{(\arabic{equation}\alph{thesubeg})}{\usecounter{thesubeg}
\leftmargin 1.2cm}}{\end{list}}
\newenvironment{eg}[1]{\refstepcounter{equation}\label{#1}\begin{list}{(\arabic{equation})}{\usecounter{thesubeg}\leftmargin 1.2cm}}{\end{list}}

\newcounter{thesubeg}
\newcounter{thesubegfoo}

\newcommand{\refegs}[2]{\setcounter{thesubegfoo}{#2}(\ref{#1}\alph{thesubegfoo})}
\newcommand{\refeg}[1]{(\ref{#1})}

\newtheorem{Exmpl}{Example}

\newcommand{\startx} 
   {\par\noindent
    \begin{minipage}[t]{5.5in}
    \vspace*{0.25ex}
    \begin{Exmpl}
    \rm \ \\
    \makebox[.1in]{}
    \begin{minipage}[t]{5in}}

\newcommand{\stopx}  
    {\end{minipage}
     \end{Exmpl}
     \vspace*{0.1ex}
     \end{minipage}\\}

\include{psfig}

\newcommand{\spud}{{\sc spud}}

\begin{document}
\begin{center}
{\bf TEXTUAL ECONOMY THROUGH CLOSE COUPLING OF SYNTAX AND SEMANTICS} \\[1em]
Matthew Stone \hspace{1em} Bonnie Webber \\
Dept. of Computer \& Information Science \\
University of Pennsylvania \\
200 South 33rd Street \\
Philadelphia PA 19104-6389 USA \\
{\tt matthew@linc.cis.upenn.edu, bonnie@central.cis.upenn.edu} \\
\end{center}

\begin{abstract}
We focus on the production of efficient descriptions of objects, actions
and events. We define a type of efficiency, {\em textual economy},
that exploits the hearer's recognition of inferential
links to material elsewhere within a sentence.
Textual economy leads to efficient descriptions because the
material that supports such inferences has been included to satisfy
independent communicative goals, and is therefore {\em overloaded}
in the sense of Pollack \cite{pollack:overloading}.
We argue that achieving textual economy imposes strong requirements on
the representation and reasoning used in generating sentences. The
{\em representation} must support the generator's simultaneous
consideration of syntax and semantics. {\em Reasoning}
must enable the generator to assess quickly and reliably at any
stage how the hearer will interpret the current sentence, with its
(incomplete) syntax and semantics. We show that these representational
and reasoning requirements are met in the \spud\ system for sentence
planning and realization.
\end{abstract}

\section{Introduction}

	The problem we address is that of producing efficient
   descriptions of objects, collections, actions, events, etc. (i.e.,
   any {\em generalized individual} from a rich ontology for Natural
   Language such as those described in \cite{bach:lectures} and advocated
   in \cite{hobbs:promiscuity}).
	We are interested in a particular kind of efficiency that we
   call {\em textual economy}, which presupposes a view of sentence
   generation as {\em goal-directed activity} that has broad support
   in Natural Language Generation (NLG) research
   \cite{appelt:planning,dale92,j.moore:phd,moore/paris:cl/planning}.
   According to this view, a system has certain communicative
   intentions that it aims to fulfill in producing a description.  For
   example, the system might have the goal of identifying an
   individual or action $\alpha$ to the hearer, or ensuring that the
   hearer knows that $\alpha$ has property {\em P}. Such goals can be
   satisfied explicitly by assembling appropriate syntactic
   constituents---for example, satisfying the goal of identifying an
   individual using a noun phrase that refers to it or identifying an
   action using a verb phrase that specifies it. {\em Textual economy}
   refers to satisfying such goals implicitly, by exploiting the
   hearer's (or reader's) recognition of inferential links to material
   elsewhere in the sentence that is there to satisfy
   independent communicative goals.
   Such material is therefore {\em overloaded}
   in the sense of \cite{pollack:overloading}.%
\footnote{
	Pollack used the term {\em overloading} to refer to cases
   where a single intention to act is used to wholly or partially
   satisfy several of an agent's goals simultaneously.
}
   While there are other ways of increasing the efficiency of
   descriptions (Section~\ref{other:sec}), our focus is on
   the efficiency to be gained by viewing a large part of generation
   in terms of describing (generalized) individuals.

   Achieving this however
   places strong requirements on the representation and reasoning
   used in generating sentences.  The {\em representation} must
   support the generator's proceeding incrementally through the syntax
   and semantics of the sentence as a whole. 
   The {\em reasoning} used
   must enable the generator to assess quickly and reliably at any
   stage how the hearer will interpret the current sentence, with its
   (incomplete) syntax and semantics.  Only by evaluating the status
   of such key questions as
\begin{itemize}
\addtolength{\itemsep}{-4pt}
\item what (generalized) individuals could the sentence (or its parts)
      refer to?
\item what (generalized) individuals would the hearer take the sentence to
      refer to?
\item what would the sentence invite the hearer to conclude about
   those individuals?
\item how can this sentence be modified or extended?
\end{itemize}
	can the generator recognize and exploit an opportunity for
   textual economy.
	
	These representational and reasoning requirements are met in
   the \spud\ system for sentence planning and realization
   \cite{colloc-paper,gen-paper}.  \spud\ draws on earlier work by
   Appelt \cite{appelt:planning} in building sentences using planning
   techniques.  \spud\ plans the syntax and semantics of a sentence by
   incorporating lexico-grammatical entries into a partial sentence
   one-by-one and incrementally assessing the answers to the questions
   given above.  In this paper, we describe the intermediate
   representations that allow \spud\ to do so, since these
   representations have been glossed over in earlier presentations
   \cite{colloc-paper,gen-paper}.  Reasoning in \spud\  is performed using
   a fast modal theorem prover \cite{fallsym97,stone:phdthesis} to
   keep track both of what the sentence {\em entails} and what the
   sentence {\em requires} in context.  By reasoning about the
   {\em predicated} relationships {\em within} clauses and the {\em
   informational} relationships \cite{moore/pollack:problem} {\em
   between} clauses, \spud\ is able to generate sentences that exhibit
   two forms of textual economy: {\em referential
   interdependency} among noun phrases within a single clause, and
   {\em pragmatic overloading} of clauses in instructions
   \cite{dieugenio/webber:overloading}.

\begin{figure}
\begin{center}
\mbox{
\psfig{figure=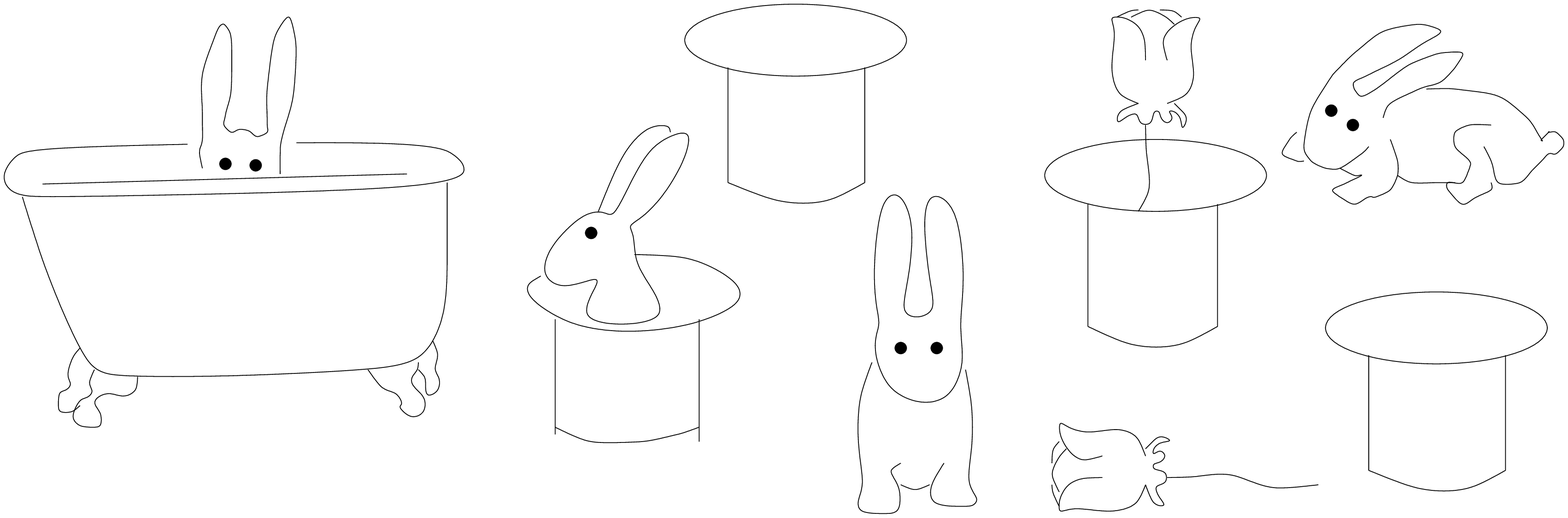,scale=30,silent=}
}
\end{center}
\vspace*{-2ex}
\caption{``Remove the rabbit from the hat.''}
\label{rabbit-out:fig}
\vspace*{-1ex}
\end{figure}

	For an informal example of the textual economy to be gained by
   taking advantage of {\em predicated} relationships within clauses,
   consider the scene pictured in Figure~\ref{rabbit-out:fig} and the
   goal of getting the hearer to take the rabbit currently in the hat
   out of the hat it's currently in.  Even though there are several
   rabbits, several hats, and even a rabbit in a bathtub and a flower
   in a hat, it would be sufficient here to issue the command:
\begin{eg}{rabbit:eg}
\item \
	{\em Remove the rabbit from the hat.}
\end{eg}
   It suffices because one of the semantic features of the verb {\em
   remove}---that its object (here, the rabbit) starts out in the
   source (here, the hat)---distinguishes the intended rabbit and hat
   in Figure~\ref{rabbit-out:fig} from the other ones.%
%

	{\em Pragmatic overloading}
   \cite{dieugenio/webber:overloading} illustrates how an
   informational relation between clauses can support textual economy
   in the clauses that serve as its ``arguments''.  In
   \cite{dieugenio/webber:overloading}, we focused on describing
   (complex) actions, showing how a clause
   interpreted as conveying the {\em goal} $\beta$ or {\em termination
   condition} $\tau$ of an action $\alpha$ partially specified in a related
   clause forms the basis of a constrained inference that provides
   additional information about $\alpha$. For example,
\begin{egs}{hold}
\item {\em Hold the cup under the spigot...}
\item {\em ...to fill it with coffee.}
\end{egs}
	Here, the two clauses \refegs{hold}{1} and \refegs{hold}{2}
   are related by purpose---specifically, enablement. The action
   $\alpha$ described in \refegs{hold}{1} will enable the actor to
   achieve the goal $\beta$ described in \refegs{hold}{2}.  While
   $\alpha$ itself does not specify the orientation of the cup under
   the spigot, its purpose $\beta$ can lead the hearer to an
   appropriate choice---to fill a cup with coffee, the cup must be
   held vertically, with its concavity pointing upwards.  As noted in
   \cite{dieugenio/webber:overloading}, this constraint depends
   crucially on the {\em purpose} for which $\alpha$ is performed.
   The purpose specified in \refegs{hold2}{2} does not constrain cup
   orientation in the same way:
\begin{egs}{hold2}
\item \
	{\em Hold the cup under the faucet...}
\item \
	{\em ...to wash it.}
\end{egs}

	Examples like \refeg{rabbit:eg} and \refeg{hold} suggest that
   the natural locality for sentence planning is in a description of a
   generalized individual.  Even though such descriptions may play out
   over several clauses (or even sentences), the predications
   within clauses and the informational relations across clauses of a
   description give rise to similar textual economies, that merit a
   similar treatment.

\section{SPUD}
\label{spud:sec}

	An NLG system must satisfy at least three constraints in
   mapping the content planned for a sentence onto the string of words
   that realize it
   \cite{dale:expressions,meteer:gap,rambow/korelsky:sp}.  Any fact to
   be communicated must be fit into an abstract grammatical structure,
   including lexical items.  Any reference to a domain entity must be
   elaborated into a description that distinguishes the entity from
   its {\em distractors}---the salient alternatives to it in context.
   Finally, a surface form must be found for this conceptual
   material.

	In one architecture for NLG systems that is becoming something
   of a standard \cite{reiter/dale:applied}, these tasks are performed
   in separate stages.  For example, to refer to a uniquely
   identifiable entity {\em x} from the common ground, first a set of
   concepts is identified that together single out {\em x} from its
   distractors in context.  Only later is the syntactic structure that
   realizes those concepts derived.

	\spud\ \cite{colloc-paper,gen-paper} integrates these
   processes in generating a description---producing both syntax and
   semantics simultaneously, in stages, as illustrated
   in~\refeg{simple-spud:eg}.
\begin{eg}{simple-spud:eg}
\item
\fbox{
\hspace*{-1em}
\mbox{
	\leaf{{\sc np}$\downarrow$:x}
	\tree
}
\psfig{figure=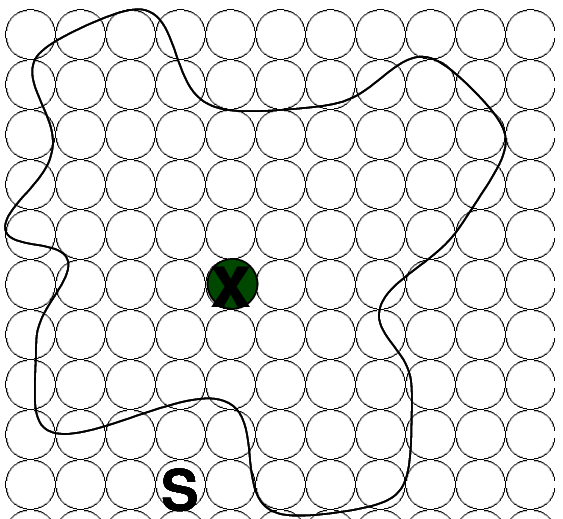,scale=33,silent=}
}
\hspace*{\fill}
\fbox{
\hspace*{-1em}
\mbox{
	\leaf{the}
	\branch{1}{\sc det}
	\leaf{book}
	\branch{1}{{\sc n}:x}
	\branch{2}{{\sc np}:x}
	\tree
}
\psfig{figure=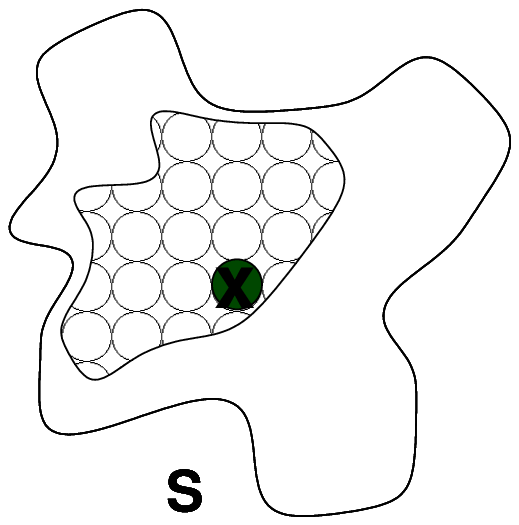,scale=33,silent=}
}
\hspace*{\fill}
\fbox{
\hspace*{-1em}
\mbox{
	\leaf{the}
	\branch{1}{\sc det}
	\leaf{tuber}
	\branch{1}{{\sc n}}
	\leaf{book}
	\branch{1}{{\sc n}:x}
	\branch{2}{{\sc n}:x}
	\branch{2}{{\sc np}:x}
	\tree
}
\psfig{figure=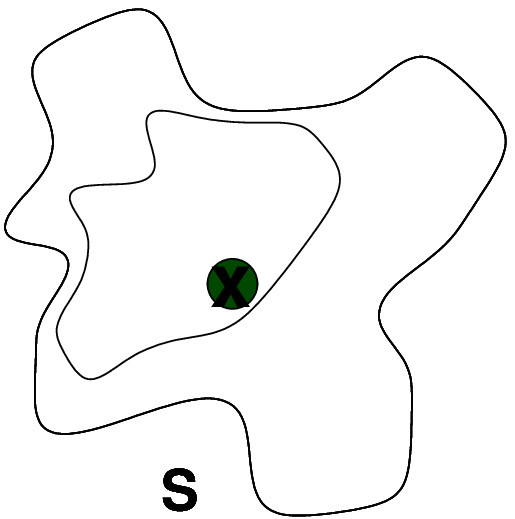,scale=33,silent=}
}
\hspace*{\fill}
\end{eg}
	Each step adds to the representation a lexicalized entry
   encoded as an elementary tree in Feature-based Lexicalized
   Tree-Adjoining Grammar (LTAG) \cite{schabes90}.  A tree may contain
   multiple lexical items (cf. ~\refeg{simple-spud:eg}b).
   Each such tree is paired with logical
   formulae that, by referring to a rich discourse model, characterize
   the semantic and pragmatic contribution that it makes to the
   sentence.  We give a detailed example of \spud 's processing in
   Section~\ref{derive:sec} and describe in Section~\ref{detail:sec}
   the reasoning methods we use to derive computational
   representations like the set of distractors shown in
   \refeg{simple-spud:eg}.  For now, a general understanding of \spud\
   suffices---this is provided by the summary in
   Figure~\ref{spud-alg:fig}.
\begin{figure}
\hspace*{\fill}\fbox{
\parbox{6in}{
\begin{itemize}
\item
	Start with a tree with one node (e.g., {\sc s}, {\sc np})
    and one or more referential or informational goals.
\item
	While the current tree is incomplete, or its references are
   ambiguous to the hearer, or its meaning does not fully convey the
   informational goals (provided progress is being made):
\begin{itemize}
\item
	consider the trees that extend the current one by the addition
   (using LTAG operations) of a true and appropriate lexicalized
   descriptor;
\item
	rank the results based on local factors (e.g., completeness of
   meaning, distractors for reference, unfilled substitution sites,
   specificity of licensing conditions); 
\item
	make the highest ranking new tree the current tree.
\end{itemize}
\vspace*{-1ex}
\end{itemize}
}~~}\hspace*{\fill}
\vspace*{-1ex}
\caption{An outline of the \spud\ algorithm}
\label{spud-alg:fig}
\end{figure}

	The procedure in Figure~\ref{spud-alg:fig} is sufficiently general
   so that \spud\ can use similar steps to construct both definite and
   indefinite referring forms.  The main
   difference lies how alternatives are evaluated.  When an indefinite
   referring form is used to refer to a {\em brand-new} generalized
   individual \cite{prince:taxonomy} (an object, for example, or an
   action in an instruction),
   the object is marked as new and does not have to be
   distinguished from others because the hearer creates a fresh ``file
   card'' for it.  However, because the domain typically provides
   features needed in an appropriate description for the object,
   \spud\ continues its incremental addition of content to convey
   them.  When an indefinite form is used to refer to an old object
   that cannot be
   distinguished from other elements of a uniquely identifiable set
   (typically an {\em inferrable} entity \cite{prince:taxonomy}),
   a process like that illustrated in
   \refeg{simple-spud:eg} must build a description that identifies
   this set, based on the known common properties of its elements.

	Several advantages of using LTAG in such an integrated system
   are described in \cite{gen-paper} (See also previous work on using TAG
   in NLG such as \cite{joshi87-gen} and \cite{yang:systemic/tag}).
   These advantages include:
\begin{itemize}
\addtolength{\itemsep}{-4pt}
\item 
	{\em Syntactic constraints can be handled early and
   naturally.}  In the problem illustrated in
   \refeg{simple-spud:eg}, \spud\ directly encodes the syntactic
   requirement that a description should have a head noun---missing
   from the concept-level account---using the {\sc np} substitution
   site.
\item
	{\em The order of adding content is flexible.}  Because an
   LTAG derivation allows modifiers to adjoin at any step (unlike a
   top-down CFG derivation), there is no tension between providing
   what the syntax requires and going beyond what the syntax requires.
\item
	{\em Grammatical knowledge is stated once only.}  All
   operations in constructing a sentence are guided by LTAG's
   lexicalized grammar; by contrast, with separate processing, the
   lexicon is split into an inventory of concepts (used for organizing
   content or constructing descriptions) and a further inventory of
   concepts in correspondence with some syntax (for surface
   realization).
\end{itemize}

	This paper delineates a less obvious, but equally significant
   advantage that follows from the ability to consider multiple goals
   in generating descriptions, using a representation and a reasoning
   process in which syntax and semantics are more closely linked:
\begin{itemize}
\addtolength{\itemsep}{-4pt}
\item {\em It naturally supports textual economy.}
\end{itemize}

\section{Achieving Textual Economy}
\label{derive:sec}

	To see how \spud\ supports textual economy, consider first how
   \spud\ might derive the instruction in Example~\refeg{rabbit:eg}.
   For simplicity, this explanation assumes \spud\ makes a
   nondeterministic choice from among available lexical entries; this
   suffices to illustrate how \spud\ can realize the textual economy
   of this example.
	
	{\em A priori}, \spud\ has a general goal of describing a new
   action that the hearer is to perform, by making sure the hearer can
   identify the key features that allow its performance.  For
   \refeg{rabbit:eg}, then, \spud\ is given two features of the action
   to be described: it involves motion of an intended object by the
   agent, and its result is achieved when the object reaches a place
   decisively away from its starting point.

	The first time through the loop of Figure~\ref{spud-alg:fig},
   \spud\ must expand an {\sc s} node.  One of the applicable moves is
   to substitute a lexical entry for the verb {\em remove}.  Of the
   elements in the verb's LTAG tree family, the one that fits the
   instructional context is the imperative tree of \refeg{remove:eg}.
\begin{egs}{remove:eg}
\item \
Syntax:
	\leaf{$\epsilon$}
        \branch{1}{{\sc np}: $\langle${\sc remover}$\rangle$}
	\leaf{remove}
	\branch{1}{{\sc v}}
	\leaf{{\sc np}$\downarrow$: $\langle${\sc removed}$\rangle$}
	\branch{2}{{\sc vp}: 
		   $\langle${\sc time,removing,source}$\rangle$}
	\branch{2}{{\sc s}:
		   $\langle${\sc time,removing}$\rangle$}
	\tree
\item \
Semantics: 
\begin{tabular}{l}
	nucleus({\sc prep}, {\sc removing}, {\sc result}) $\wedge$
	in({\sc prep}, start({\sc time}), {\sc removed}, {\sc source}) $\wedge$ \\
        caused-motion({\sc removing}, {\sc remover}, {\sc removed}) $\wedge$ \\
	away({\sc result}, end({\sc time}), {\sc removed}, {\sc source})
\end{tabular}
\end{egs}
	The tree given in \refegs{remove:eg}{1} specifies that {\em
   remove} syntactically {\it satisfies} a requirement to include an
   {\sc s}, {\it requires} a further {\sc np} to be included
   (describing what is removed), and {\it allows} the possibility of
   an explicit {\sc vp} modifier that describes what the latter has
   been removed from.%
\footnote{
	Other possibilities are that {\sc source} is not mentioned
   explicitly, but is rather inferred from (1) the previous discourse
   or, as we will discuss later, (2) either the
   predicated relationships within the clause or its informational
   relationship to another clause.
}
	The semantics in \refegs{remove:eg}{2} consists of a set of
   features, formulated in an ontologically promiscuous semantics, as
   advocated in \cite{hobbs:promiscuity}.  It follows
   \cite{moens/steedman:ontology} in viewing events as consisting of a
   preparatory phase, a transition, and a result state (what is called
   a {\em nucleus} in \cite{moens/steedman:ontology}).  The semantics
   in \refegs{remove:eg}{2} describes all parts of a {\em remove}
   event: In the preparatory phase, the object ({\sc removed}) is
   in/on {\sc source}.  It undergoes motion caused by the agent ({\sc
   remover}), and ends up away from {\sc source} in the result state.

	Semantic features are used by \spud\ in one of two ways.
   Some make a {\em semantic contribution} that specifies new
   information---these add to what new information the speaker can
   convey with the structure.  Others simply impose a {\em semantic
   requirement} that a fact must be part of the conversational
   record---these figure in ruling out distractors.  

	For this instruction, \spud\ treats the {\sc caused-motion}
   and {\sc away} semantic features as semantic contributions. It
   therefore determines that the use of this item communicates the
   needed features of the action.  At the same time, it treats the
   {\sc in} feature---because it refers to the shared initial state in
   which the instruction will be executed---and the {\sc nucleus}
   feature---because it simply refers to our general ontology---as
   semantic requirements.  \spud\ therefore determines that the only
   $\langle${\sc removed},{\sc source}$\rangle$ pairs that the hearer
   might think the instruction could refer to are pairs where {\sc
   removed} starts out in/on {\sc source} as the action begins.

	Thus, \spud\ derives a triple effect from use of the word {\em
   remove}---increasing {\em syntactic satisfaction}, making {\em
   semantic contributions} and satisfying {\em semantic
   requirements}---all of which contribute to \spud's task of
   completing an {\sc s} syntactic constituent that conveys needed
   content and refers successfully.  Such multiple effects make it
   natural for \spud\ to achieve textual economy.  Positive effects on
   any of the above dimensions can suffice to merit inclusion of an
   item in a given sentence.  However, the effects of inclusion may go
   beyond this: even if an item is chosen for its semantic
   contribution, its semantic requirements can still be exploited in
   establishing whether the current lexico-syntactic description is
   sufficient to identify an entity, and its syntactic contributions
   can still be exploited to add further content.

   	Since the current tree is incomplete and referentially
   ambiguous, \spud\ repeats the loop of Figure~\ref{spud-alg:fig},
   considering trees that extend it.  One option is to
   adjoin at the {\sc vp} the entry corresponding to {\em from the
   hat}.  In this compound entry, {\em from} matches the verb and {\em
   the} matches the context; {\em hat} carries semantics, requiring
   that {\sc source} be a hat.  After adjunction, the requirements
   reflect both {\em remove} and {\em hat}; reference, \spud\
   computes, has been narrowed to the hats that have something in/on
   them (the rabbit, the flower).

	Another option is to substitute the entry for {\em the rabbit}
   at the object {\sc np}; this imposes the requirement that {\sc
   removed} be a rabbit.  Suppose \spud\ discards this option in this
   iteration, making the other (perhaps less referentially ambiguous)
   choice.  At the next iteration, {\em the rabbit} still remains an
   option.  Now combining with {\em remove} and {\em hat}, it derives
   a sentence that \spud\ recognizes to be complete and referentially
   unambiguous, and to satisfy the informational goals.
   
   	Now we consider the derivation of \refeg{hold}, which shows
   how an informational relation between clauses can support textual
   economy in the clauses that serve as its ``arguments''.  \spud\
   starts with the goal of describing the holding action in the main
   clause, and (if possible) also describing the filling action and
   indicating the purpose relation (i.e., {\em enablement}) between
   them.  For the {\em holding} action, \spud's goals include making
   sure that the sentence communicates where the cup will be held and
   how it will be held (i.e., {\sc upward}).  \spud\ first selects an
   appropriate lexico-syntactic tree for imperative {\em hold}; \spud\
   can choose to adjoin in the purpose clause next, and then to
   substitute in an appropriate lexico-syntactic tree for {\em fill}.
   After this substitution, the semantic contributions of the sentence
   describe an action of {\em holding an object} which {\em generates}
   an action of {\em filling that object}.  As shown in
   \cite{dieugenio/webber:overloading}, these are the premises of an
   inference that the object is held upright during the filling.  When
   \spud\ queries its goals at this stage, it thus finds that it has
   in fact conveyed how the cup is to be held.  \spud\ has no reason
   to describe the orientation of the cup with additional content.

   	Additional examples of using \spud\ to generate instructions
   can be found in \cite{bourne:phd,stone:phdthesis}.

\section{Assessing interpretation in \spud}
\label{detail:sec}

	This section describes in a bit more detail how \spud\
   computes the effects of incorporating a particular lexical item
   into the sentence being constructed.  For a more extensive
   discussion, see \cite{stone:phdthesis}.

	\spud's computations depend on its representation of overall
   contextual background, including the status of propositions
   and entities in the discourse.  For the purpose of generating
   instructions to a single hearer, we assume
   that any {\em proposition} falls either within the private
   knowledge of the speaker or within the common ground that speaker
   and hearer share.  We implement this distinction by specifying
   facts in a modal logic with an explicit representation of
   knowledge: $\speaker\mbox{\em p}$ means that the speaker knows {\em
   p}; $\shared\mbox{\em p}$ means that {\em p} is part of the common
   ground.  Each {\em entity}, $\entity$, comes with a context set
   $\distractors(\entity)$ including it and its distractors.
   Linguistically, when we have $\entityA \in \distractors(\entityB)$
   but not $\entityB \in \distractors(\entityA)$, then $\entityA$ is
   more salient than $\entityB$.
	
	This conversational background serves as a resource for
   constructing and evaluating a three-part state-record for an
   incomplete sentence, consisting of:
\begin{itemize}
\addtolength{\itemsep}{-4pt}
\item
	An {\em instantiated} tree describing the syntactic structure
   of the sentence under construction.  Its nodes are labeled by a
   sequence of variables $\variables$ indicating the patterns of
   coreference in the tree; but the tree also records that the speaker
   intends $\variables$ to refer to a particular sequence of
   generalized individuals $\entities$.
\item
	The {\em semantic requirements} of the tree, represented by a
   formula $\requirements(\variables)$.  This formula must match facts
   in the common ground; in our modal specification, such a match
   corresponds to a proof whose conclusion instantiates $\shared
   \requirements(\variables)$.  In particular, the speaker ensures
   that such a proof is available when $\variables$ is instantiated to
   the entities $\entities$ that the speaker means to refer to.  This
   determines what alternative referents that the hearer may still
   consider: $\{ \; \entitiesA \in \distractors(\entities) \; | \;
   \shared \requirements(\entitiesA) \; \}$.  The semantic
   requirements of the tree result from conjoining the requirements
   $\requirementsI(\variablesI)$ of the individual lexical items from
   which the state is derived.
\item
   	The {\em semantic contributions} of the tree, represented by a
   formula $\contributions(\variables)$; again, this is the
   conjunction of the contributions $\contributionsI(\variablesI)$ of
   the individual items.  These contributions are added to the
   common ground, allowing both speaker and hearer to draw shared
   conclusions from them.  This has inspired the following test for whether
   a goal to communicate $\commgoal$ has been indirectly achieved. Consider
   the content of the discourse as represented by $\shared$, augmented
   by what this sentence will contribute (assuming we identify
   entities as needed for reference): $\contributions(\entities)$.
   Then if $\commgoal$ follows, the speaker has conveyed what is
   needed.  
\end{itemize}

	When \spud\ considers extending a state by a lexical item, it
   must be able to update each of these records quickly.  The heart of
   \spud's approach is logic programming~\cite{stone:phdthesis}, which
   links complexity of computation and complexity of the domain in a
   predictable way.  For example, informational goals are assessed by
   the query $\shared(\contributions(\entities) \imp \commgoal)$.
   This leaves room for inference when necessary, without swamping
   \spud; in practice, $\commgoal$ is often a primitive feature of the
   domain and the query reduces to a simple matching operation.
   Another source of tractability comes from combining logic
   programming with special-purpose reasoning.  For example, in
   computing reference, $\{ \; \entitiesIA \in
   \distractors(\entitiesI) \; | \; \shared
   \requirementsI(\entitiesIA) \; \}$ is found using logic programming
   but the overall set of alternatives is maintained using
   arc-consistency constraint-satisfaction, as in
   \cite{dale/haddock:referring,haddock:thesis}.

	\spud\  must also settle which semantic features are taken to
   constitute the semantic {\em requirements} of the lexical item and
   which are taken to constitute its semantic
{\em contributions}.\footnote{
	These can vary with context: consider a variant on Figure
   \ref{rabbit-out:fig}, where the hearer is asked ``What just
   happened?''.
\begin{center}
\mbox{
\psfig{figure=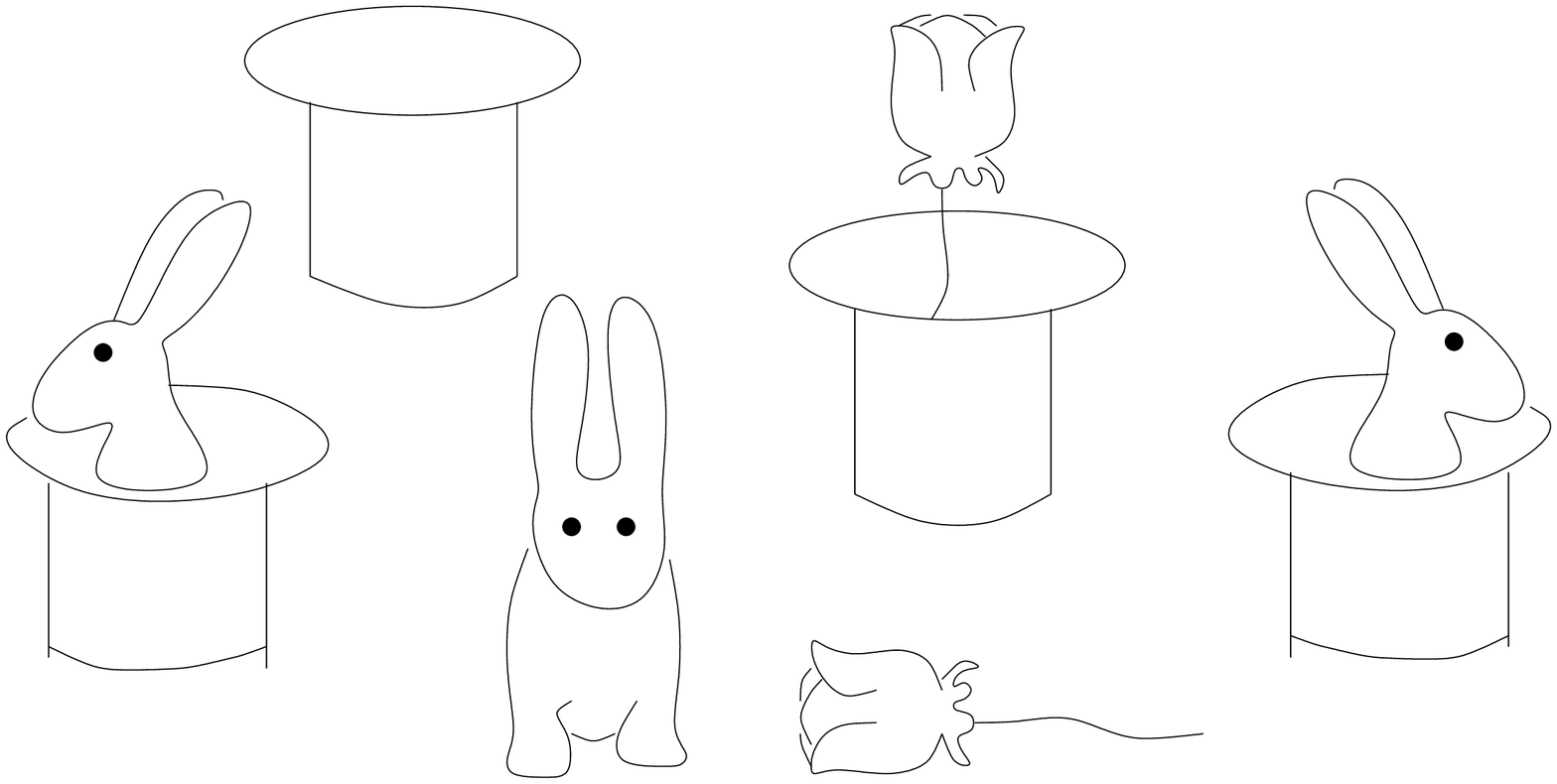,scale=25,silent=}
}
\end{center}
   One possible response --- ``I have removed the rabbit from the
   hat'' --- refers successfully, despite the many rabbits and hats,
   because there is still only one rabbit in this scene that could
   have been removed from a hat.  Here, where the scene is taken as
   shared, what is taken as a semantic requirement of {\em
   remove}---that the rabbit ends up away from the hat---is used to
   identify a unique rabbit.  This contrasts with the previous
   ``rabbit'' example where, taking the scene in Figure
   \ref{rabbit-out:fig} as shared, the command ``Remove the rabbit
   from the hat'' takes as its semantic requirement that the rabbit be
   in the hat and uses it for unique identification.  Note that if
   the above scene is not taken as shared, both are then taken as
   semantic contributions, and ``I have removed a rabbit from a hat''
   becomes an acceptable answer.
}
	When \spud\ partitions the semantic features of the lexical
   item, as many features as possible are cast as requirements---that
   is, the item links as strongly with the context as possible.  In
   some cases, the syntactic environment may further constrain this
   assignment.  For example, we constrain items included in a definite
   {\sc np} to be semantic {\em requirements}, while the main verb in
   an indicative sentence is usually taken to make a semantic {\em
   contribution}. (Exceptions to such a policy are justified in
   \cite{walker:thesis}.)

\section{Other Methods that Contribute to Efficient Descriptions}
\label{other:sec}

   	This section contrasts \spud---and its close coupling of
   syntax and semantics---with prior work on generating more concise
   descriptions by considering the effects of broader goals,%
\footnote{
	Other ways of making descriptions more concise, such as
   through the use of anaphoric and deictic pronouns (or even
   pointing, in multi-modal contexts), are parasitic on the hearer's
   focus of attention, which can (in large part) be defined
   independently of goal-directed features of text.
}
  	starting with Appelt \cite{appelt:planning}.  Appelt's planning
   formalism includes plan-critics that can detect and collapse
   redundancies in sentence plans.  However, his framework treats
   subproblems in generation as independent by default; and writing
   tractable and general critics is hampered by the absence of
   abstractions like those used in \spud\ to simultaneously model the
   syntax and the interpretation of a whole sentence.

   	\cite{dale/haddock:referring,horacek:referring,mcdonald:redundancy},
   in contrast, use specialized mechanisms to capture particular
   descriptive efficiencies.  By using syntax to work on inferential
   and referential problems simultaneously, \spud\ captures such
   efficiencies in a uniform procedure.  For example, in
   \cite{mcdonald:redundancy}, McDonald considers descriptions of
   events in domains which impose strong constraints on what
   information about events is semantically relevant.  He shows that
   such material should and can be omitted, if it is both
   syntactically optional and inferentially derivable:
\begin{quote}
{\sc fairchild} Corporation (Chantilly VA) Donald E Miller was named
senior vice president and general counsel, succeeding Dominic A Petito,
who resigned in November, at this aerospace business. Mr. Miller, 43 years
old, was previously principal attorney for Temkin \& Miller Ltd.,
Providence RI.
\end{quote}
	Here, McDonald points out that one does not need to explicitly
   mention the position that Petito resigned from in specifying the
   resignation sub-event, since it must be the same as the one that
   Miller has been appointed to.  This can be seen as a special case
   of pragmatic overloading.

\begin{figure}
\centerline{
\mbox{
\psfig{figure=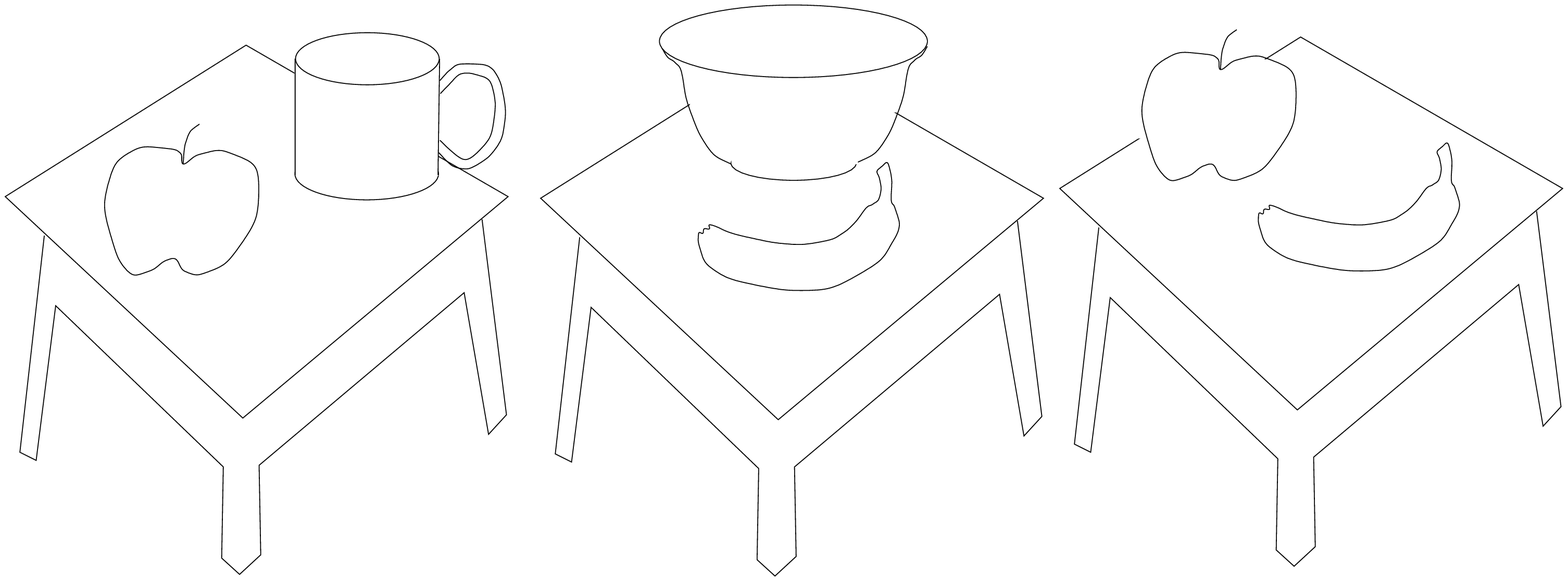,scale=28,silent=}
}}
\vspace*{-2ex}
\caption{``The table with the apple and with the banana''}
\label{banana:fig}
\end{figure}
	Meanwhile, Dale and Haddock \cite{dale/haddock:referring}
   consider generating interacting references, building on Haddock's
   work on reference resolution \cite{haddock:thesis}.  Their example
   {\sc np}, {\em the rabbit in the hat}, refers successfully in a
   context with many rabbits and many hats, so long as only one of the
   rabbits, $\mbox{\em r}_5$ say, is actually in one of the hats,
   $\mbox{\em h}_3$ say.  Like~\refeg{rabbit:eg}, the efficiency of
   this description comes from the uniqueness of this rabbit-hat pair.
   However, Dale and Haddock construct {\sc np} semantics in isolation
   and adopt a fixed, depth-first strategy for adding content.
   Horacek \cite{horacek:referring}, challenges this strategy with
   examples that show the need for modification at multiple points in
   an {\sc np}.  For example, \refeg{Horacek:eg} refers with respect
   to the scene in Figure~\ref{banana:fig}.
\begin{eg}{Horacek:eg}
\item \
	{\em the table with the apple and with the banana}.
\end{eg}
	\refeg{Horacek:eg} identifies a unique table by exploiting its
   association with two objects it supports: the apple and the banana
   that are on it.  (Note the other tables, apples and bananas in the
   figure---and even tables with apples and tables with bananas.)
   Reference to one of these---the apple, say---is incorporated into
   the description first; then that (subordinate) entity is identified
   by further describing the table (higher up).%
\footnote{
	Horacek also stresses that there is no need to describe these
   auxiliary objects separately or identify them uniquely.  For
   example, {\em the table with two fruits} provides a possible
   alternative to \refeg{Horacek:eg}.  We see no principled obstacle
   to expressing the same insight in \spud.  However, because of
   \spud's close coupling between syntax and semantics, this analysis
   must await development of a semantic analysis of plurality in
   \spud.
}
   By considering sentences rather than isolated noun phrases, \spud\
   extends such descriptive capacities even further.
	
\section{Remarks and Conclusion}

	In this paper, we have shown how the semantics associated with
   predication within clauses and informational relations between
   clauses can be used to achieve textual economy in a system (\spud)
   that closely couples syntax and semantics.  In both cases,
   efficiency depends only on the informational consequences of
   current lexico-syntactic choices in describing the {\em
   generalized} individual of interest; there is no appeal to information
   available in the discourse context, which is already well-known as
   a source of economy, licensing the use of anaphoric and deictic
   forms, the use of ellipsis, etc. Thus, we claim that this approach
   truly advances current capabilities in NLG.

	Finally, we must make clear that we are talking about the {\em
   possibility} of producing a particular description (one in which a
   wider range of inferrable material is elided); we are {\em not}
   making claims about a particular algorithm that exploits such a
   capability. Thus it is not relevant here to question computational
   complexity or look for a comparison with algorithms previously
   proposed by Dale, Reiter, Horacek and others
   \cite{dale:expressions,horacek:referring,reiter:nouns} that compute
   ``minimal'' descriptions of some form.  Currently, the control
   algorithm used in the \spud\ generator is the simple greedy
   algorithm described in \cite{colloc-paper,gen-paper} and summarized
   in Figure~\ref{spud-alg:fig}.  The important point is that the
   process enables inferences to be performed that allow more
   economical texts: the next step is to address the complexity issues
   that these other authors have elaborated and show how \spud's
   description extension and verification process can be incorporated
   into a more efficient or more flexible control structure.

\section{Acknowledgments}
	Support for this work has come from an IRCS graduate
   fellowship, the Advanced Research Project Agency (ARPA) under grant
   N6600194C6-043 and the Army Research Organization (ARO) under grant
   DAAHO494GO426.  Thanks go to Mark Steedman and members of the SPUD
   group: Christy Doran, Gann Bierner, Tonia Bleam, Julie Bourne,
   Aravind Joshi, Martha Palmer, and Anoop Sarkar.

\end{document}